\pdfoutput=1
\documentclass[letterpaper, 10 pt, conference]{ieeeconf}  

\IEEEoverridecommandlockouts                              

\usepackage[hidelinks]{hyperref}
\usepackage[font=small,skip=10pt]{caption}
\usepackage{graphics} 
\usepackage{graphicx}
\usepackage{algpseudocode}
\usepackage{algorithm}
\usepackage[]{units}
\usepackage{pgfplots}
\usepackage{tikz}
\usetikzlibrary{positioning}
\usetikzlibrary{shapes.geometric}
\usetikzlibrary{shapes.misc}
\usetikzlibrary{external}
\usetikzlibrary{arrows}

\usepackage{paralist}
\usepackage{mathtools}

\usepackage{xcolor}
\usepackage{siunitx}
\usepackage{multirow}
\usepackage{epstopdf}
\usepackage{todonotes}
\graphicspath{ {Pictures/} }
\usepackage{epstopdf}
\usepackage{tikz}
\usetikzlibrary{shapes,arrows,patterns}
\usetikzlibrary{arrows.meta}
\usetikzlibrary{plotmarks}
\usetikzlibrary{calc,decorations.markings,math}
  \usepgfplotslibrary{patchplots}
\usepackage{epsfig}
\usepackage{float}
\usepackage{amsmath} 
\usepackage{amssymb}  
\usepackage{mathtools}
\graphicspath{ {Pictures/} }
\DeclareGraphicsExtensions{.pdf,.eps}
\usepackage[acronym,nomain,nonumberlist]{glossaries}
\newacronym{mav}{MAV}{Micro Aerial Vehicle}
\newacronym{dl}{DL}{Deep Learning}
\newacronym{nmpc}{NMPC}{Nonlinear Model Predictive Control}
\newacronym{imu}{IMU}{Inertial Measurement Unit}
\newacronym{cnn}{CNN}{Convolutional Neural Network}
\newacronym{relu}{ReLU}{Rectified Linear Unit}
\newacronym{mlp}{MLP}{MultiLayer Perceptron}
\newacronym{mpc}{MPC}{Model Predictive Control}
\newacronym{nn}{NN}{Nueral Network}
\newacronym{gnss}{GNSS}{Global Navigation Satellite System}
\newacronym{sfm}{SfM}{Structure from Motion}
\newacronym{ar}{AR}{Augmented Reality}
\newacronym{ros}{ROS}{Robot Operating System}
\newacronym{panoc}{PANOC}{Proximal Averaged Newton-type method for Optimal Control}
\newacronym{gps}{GPS}{Global Positioning System}
\newacronym{vi}{VI}{Visual Inertia}
\newacronym{fps}{fps}{Frame Per Second}
\newacronym{mae}{MAE}{mean absolute error}
\newacronym{dof}{DoF}{Degree of Freedom}
\newacronym{ekf}{EKF}{Extended Kalman Filter}
\newacronym{ahe}{AHE}{Adaptive Histogram Equalization}
\newacronym{api}{API}{Application Programming Interface}
\newacronym{zepa}{ZEPA}{Zero Entry Production Areas}

\pgfplotsset{compat=1.14}
\usepackage{url}

\title{\LARGE \bf MAV Navigation in Unknown Dark Underground Mines Using Deep Learning*\thanks{*This work has been partially funded by the European Unions Horizon 2020 Research and Innovation Programme under the Grant Agreement No. 730302 SIMS.}}

\author{Sina Sharif Mansouri$^{1}$\thanks{$^{1}$ Authors contributed equally}, Christoforos Kanellakis$^{1}$, Petros Karvelis, Dariusz Kominiak, \\and George Nikolakopoulos \thanks{$^{2}$ Robotics and AI Team, Department of Computer, Electrical and Space Engineering, Lule\r{a} University of Technology, Lule\r{a} SE-97187, Sweden,  Emails:\texttt{\{sinsha, chrkan, petkar, darkom, geonik\}@ltu.se}}
}

\begin{document}
\maketitle
\thispagestyle{empty}
\pagestyle{empty}

\begin{abstract}
This article proposes a \gls{dl} method to enable fully autonomous flights for low-cost \glspl{mav} in unknown dark underground mine tunnels. This kind of environments pose multiple challenges including lack of illumination, narrow passages, wind gusts and dust. The proposed method does not require accurate pose estimation and considers the flying platform as a floating object. The \gls{cnn} supervised image classifier method corrects the heading of the \gls{mav} towards the center of the mine tunnel by processing the image frames from a single on-board camera, while the platform navigates at constant altitude and desired velocity references. Moreover, the output of the \gls{cnn} module can be used from the operator as means of collision prediction information. The efficiency of the proposed method has been successfully experimentally evaluated in multiple field trials in an underground mine in Sweden, demonstrating the capability of the proposed method in different areas and illumination levels.  
\end{abstract}

\glsresetall
\section{Introduction}
\gls{mav} have the potential to operate in a wide range of applications, such as underground mine inspection~\cite{kanellakis2018towards}, subterranean exploration~\cite{rogers2017distributed}, large infrastructure inspection~\cite{mansouri2018cooperative}, aerial terrain mapping~\cite{mansouri20182d}, and Mars canyon exploration~\cite{matthaei2013swarm}, \glspl{mav} have a potential to provide leading solutions. In these harsh, inaccessible and dangerous environments, the \glspl{mav} can grant access for the monitoring of working personnel, exploring the area and minimizing service times.  

\begin{figure}[htbp!]
  \centering
    \includegraphics[width=0.8\linewidth]{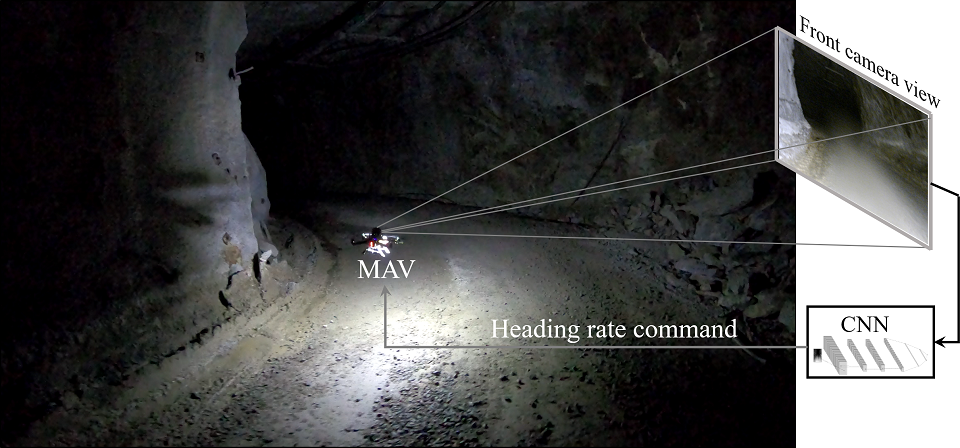}
      \caption{The proposed approach for classifying the images from the forward looking camera during autonomous navigation in underground mines. \textit{Supplementary Video:} \underline{https://youtu.be/j3N8ij9MfSA}.}
        \label{fig:mineboliden}
\end{figure}

Autonomous navigation of \glspl{mav} in underground tunnels is a challenging task. Figure~\ref{fig:mineboliden} depicts few of these challenges, such as lack of natural illumination and uneven surfaces in one part of the visited underground mine, while at the same time, these dark environments do not provide sufficient information for cameras for accurate pose estimation~\cite{ozaslan2017autonomous}.

This article proposes a \gls{dl} method for autonomous navigation of a low cost and resource-constrained \gls{mav} in underground mines. The proposed method considers a lack of information on full pose estimation and establishes the \gls{mav} as a floating object, which navigates by $x$, $y$ velocities and altitude commands. In order to correct the \gls{mav} heading and avoid collisions, the on-board forward looking camera image stream is fed to the \gls{cnn}~\cite{krizhevsky2012imagenet} module that is the most prominent member of the \gls{dl} family. To reduce noise and computation time, the image stream is re-sized to $128 \times 128$ and converted to gray scale. The \gls{cnn} classifies the images into three categories of \textit{left}, \textit{center}, and \textit{right}, where each category corresponds to a specific heading rate command for the \gls{mav}. This categorization results into preventing potential collisions to the walls, while the information from the \gls{cnn} can be used as a collision prediction warning to the human operator, being visualized through an \gls{ar} module. The efficiency of the method is evaluated in an underground mine in Sweden with different illumination levels, visual sensors and velocities of navigation.

%
\subsection{Background \& Motivation}
%
Recently, there has been an emerging effort towards developing aerial robots with capability to operate in underground environments ~\cite{schmid2014autonomous,gohl2014towards,ozaslan2017autonomous}. In~\cite{schmid2014autonomous}, a visual inertial navigation framework has been proposed, to implement position tracking control of the platform. In this case, the \gls{mav} was controlled to follow obstacle free paths, while the system was experimentally evaluated in a real scale tunnel environment. In~\cite{gohl2014towards} a \gls{vi} sensor and a laser scanner were installed on a hexacopter and the platform was manually guided across a vertical mine shaft to collect data for post-processing. The extracted information from the measurements have been utilized to create a 3D mesh of the environment and localize the vehicle. In~\cite{ozaslan2017autonomous}, the authors addressed the problem of estimation, control, navigation and mapping, for autonomous inspection of tunnels, using aerial vehicles equipped with a high-end sensor suit and the overall approach was validated through field trials. In~\cite{mansouri2019autonomouscontour}, the low cost platform is proposed for autonomous navigation in dark underground mine, however for effectiveness of the method there should be no external illumination except the light source on the \gls{mav}, this limits the generalization of the method.

Furthermore, there are few works using machine learning techniques for the problem of navigation at in-door and out-door environments, mainly due to the fact that these methods require a large amount of data and a high computation power for training in most cases a \gls{cnn}, which is an off-line procedure. However, after the training, the \gls{cnn} can be used for enabling an autonomous navigation with much lower computation power, especially when compared to the training phase. The works using \gls{cnn} for navigation, such as~\cite{adhikari2018accurate},~\cite{giusti2016machine},~\cite{ran2017convolutional}, \cite{smolyanskiy2017toward},~\cite{kourislearning}, utilized the image frame of on-board camera to feed the \gls{cnn} for providing heading commands to the platform. These works have been evaluated and tuned in out-door environments and with a good illumination with the camera, thus providing rich data about the surrounding of the platforms. Furthermore, preliminary and limited studies of \gls{mav} navigation in an underground mine using \gls{cnn} was presented in~\cite{SinaRobio2018}, however the method was evaluated in off-line collected data-sets from two underground tunnels, without the \gls{mav} in the loop.

\subsection{Contributions}


Based on the aforementioned state of the art, the main contributions of this article are provided in this section. The first and major contribution of this work is the development of a \gls{cnn} supervised image classifier to align the heading direction of the \gls{mav} along the tunnel axis, using the image stream from the on-board forward looking camera. The \gls{cnn} method identifies both the surrounding walls of the tunnel, as well as open space, acting as a collision prediction module for collision free navigation of the \gls{mav}, a concept that has not been evaluated in dark tunnels. Moreover, the \gls{mav} is operated as a floating object, and the \gls{cnn} part enables fully autonomous navigation of low-cost aerial platforms in challenging dark underground environments.

The second contribution stems from the deployment of the proposed approach in extended field trials in an underground mine $\unit[790]{m}$ deep, without natural illumination. The successful experimental results demonstrate that the developed system brings closer the establishment of autonomous \glspl{mav} in the operation cycles of underground mines. The following link \url{https://youtu.be/j3N8ij9MfSA} provides a video summary of the system.

The final contribution stems from providing to the research community open access to all collected data-sets and by that enabling further developments towards the envisioned autonomous flying in the dark tunnels.

\subsection{Outline}

The rest of the article is structured as follows. Initially, Section~\ref{Problem Statement} presents the problem formulation of the proposed method. Then, the \gls{cnn} implementation is presented in Section~\ref{sec:cnn}, followed by the evaluation results of the proposed method in an underground mine in Section~\ref{Results}. Finally, Section~\ref{Conclusions} concludes the article by summarizing the findings and offering some directions for future research.
%
\section{Problem Statement} \label{Problem Statement}
%
Integration of the \glspl{mav} in inspection and production areas is emerging in the mining industry. In these cases, the inspection task includes the deployment of the \gls{mav} equipped with sensor suites to autonomously navigate along the tunnel and collect information, such as images, gas and dust levels, 3D models, etc. The collected data will be used for further analysis to determine the status of the inspected area. This technology creates a safer working environment with increased production, while reducing the overall operation costs, which is aligned with the envisioned mine of \gls{zepa}~\cite{NIKOLAKOPOULOS201566}.

Usually, to increase the level of autonomy and provide stability and reliability of the operation, the platforms are built with high-end and expensive components. However, the long-term operation of these platforms, in harsh environments, degrades their performance and integrity over time. Thus, the objective of this article is to propose a low-cost solution for the \gls{mav} navigation in underground mine environments, while performing the task equally reliable. These platforms can be considered as consumables that can be instantly replaced. This article presents the \gls{dl} method for correcting the heading direction of the \gls{mav} without relying on accurate localization schemes. 
The platform is flying at constant altitude and potential field~\cite{kanellakis2018towards} from 2D lidar ranges $R$ generates linear velocities, while the \gls{cnn} module provides heading rate commands for avoiding collisions based on the image stream.


\begin{figure*}[htbp!] \centering
    \includegraphics[width=\linewidth]{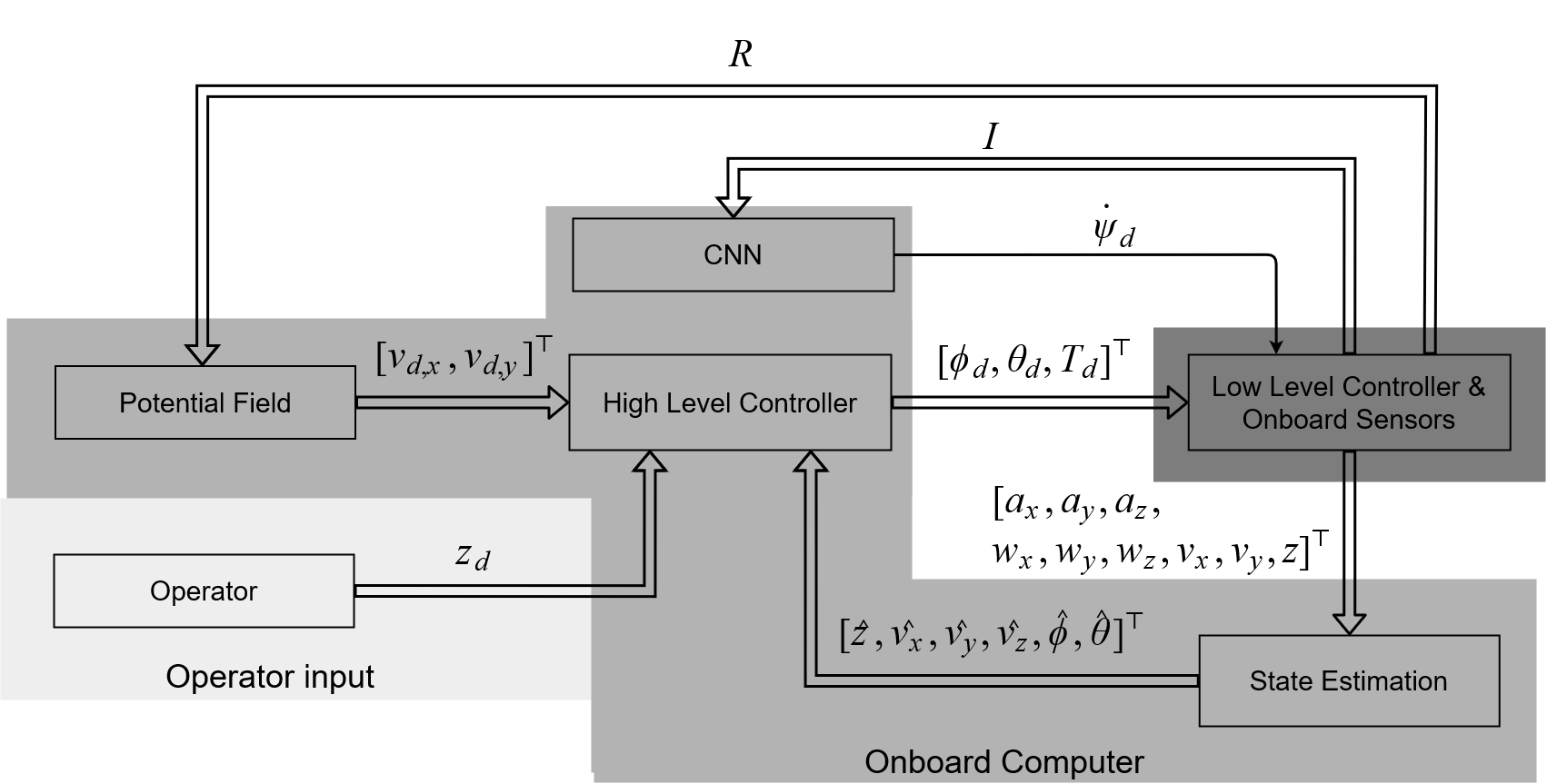}
\caption{Control scheme of the proposed method. The heading commands \gls{cnn} based approaches is used. The high level controller generates thrust and attitude commands, while the low level controller generates motor commands. The velocity estimation is based on \gls{imu}, optical flow, and one beam lidar.}
\label{fig:Controllerscheme}
\end{figure*}

Figure~\ref{fig:Controllerscheme} depicts the proposed control architecture. The \gls{mav} is presented as a floating object without information on its position on $x$ and $y$ axes, while the estimation of the altitude $z$, velocities on $x$, $y$, and $z$ axes $v_x,v_y,v_z$ and attitudes $\phi,\theta$ are obtained from \gls{imu} measurements, optical flow, one beam lidar. The desired altitude $z_{d}$ and desired velocities $v_{d,x}$, $v_{d,y}$ are fed to a high level controller to generate thrust and attitude commands $[T_d, \phi_d, \theta_d]^\top$ for the low level controller. Moreover, obtaining an accurate heading angle, especially for a low-cost navigation systems, is not always possible, since most of the methods rely on information on magnetometer and gyrocompass, multi-antenna \glspl{gnss} or position information of the platform~\cite{gade2016seven}. However, a magnetometer is not a reliable sensor especially in an underground mine and \gls{gnss} is not available for underground areas. Thus, the heading rate $\dot{\psi}$ is generated by the \gls{cnn} module. 

\section{Convolutional Neural Network} \label{sec:cnn}
Image classification can be considered as one of the fundamentals research fields in the area of Computer Vision. Among the various neural networks proposed \glspl{cnn} have played a crucial role. \glspl{cnn} are made up of a large number of layers, and each each layer is consisted by a number of neurons which have weights and biases. The input to the implemented \gls{cnn} is a fixed-size image and the output is a vector of representing the probabilities for each one of the classes that we need to detect. 

Typically the \gls{cnn}~\cite{krizhevsky2012imagenet,cirecsan2012multi} is composed of a series of non-linear processing layers namely the convolutional layer and the pooling layer which are fully connected. The convolutional layer work on the local volumes of data called patches through convolutional kernels which are called filters in order to extract a number of features from the patches. The pooling layer reduces the size of the original image and the subsequent feature maps and thus providing translation in-variance. The nonlinear activation layers (usually consisted of \glspl{relu} that have almost completely substituted the traditional sigmoid activation functions), as in the case of the conventional \glspl{nn} allows to the \glspl{cnn} to learn non-linear mappings between the input and the output. The final layers of a \gls{cnn} are ``flat'' fully connected layers identical to the ones used in conventional \glspl{mlp}. A \gls{cnn} is trained ``end to end'' to learn to map the inputs to their corresponding targets by using gradient-descent based learning rules. Same \gls{cnn} structure as presented in~\cite{SinaRobio2018} is chosen for this work.



The input matrix of the \gls{cnn} is a matrix of $128 \times 128 \time 1$ and the output is three classes of each image. The class of the image can be one from \textit{Left}, \textit{Center}, \textit{Right}. Each label represents the direction of the platform heading, e.g. in case of \textit{Center} the heading rate should be zero and in case of \textit{Left} the heading rate should have negative value to avoid the left wall. To reduce the computation time and size of data, independent to the type of the on on-board camera, the image is resized to $128 \times 128$ pixels. Furthermore, the images from the camera are converted to gray-scale because the RGB sensors do not yield any extra information about the dark environments, while the object recognition based on gray-scale images can outperform RGB image based recognition~\cite{bui2016using}, and gray-scale images reduce the overall computation time and noise. The \gls{cnn} has been implemented in Python by Keras~\cite{chollet2015keras} as a high-level neural network \gls{api}. The loss function is categorical crossentropy and the optimization is Adam optimizer~\cite{kingma2014adam}. Finally a workstation has been utilized, equipped with an Nvidia GTX 1070 GPU for the training of the network with 25 epochs and 200 steps per epoch. 

\section{Results} \label{Results}
This section describes the experimental setup and experimental evaluation of the proposed novel \gls{cnn} method in the underground mining environment. Link: \url{https://youtu.be/j3N8ij9MfSA} provides a video summary of the results.  

\subsection{Experimental Setup}
In this work two separate platforms are tested. The first platform is a quad-copter, which was developed at Lule{\aa} University of Technology based on the ROSflight~\cite{jackson2016rosflight} flight controller and the second one is the commercially available quad-copter Parrot Bebop 2~\cite{parrot2016parrot}. 

\begin{figure}[htbp!]
    \centering
 \includegraphics[width=\linewidth]{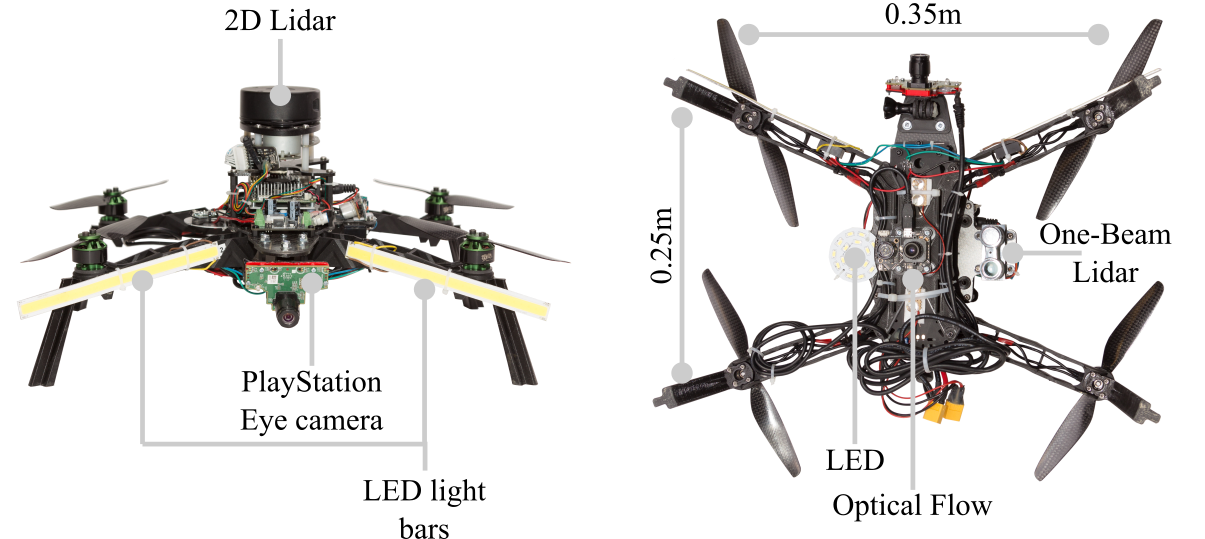}
    \caption{The developed ROSflight based quad-copter equipped.}
    \label{fig:LTUplatfrom}
\end{figure}

The ROSflight Based Quad-copter weights $\unit[1.5]{kg}$ and provides $\unit[8]{mins}$ of flight time with 4-cell~\unit[1.5]{hA} LiPo battery. The Aaeon UP-Board is the main processing unit, incorporating an Intel Atom x5-Z8350 processor and $\unit[4]{GB}$ RAM. The operating system running on the board is the Ubuntu Desktop 16.04, while \gls{ros}~\cite{quigley2009ros} Kinetic has been also included. The platform is equipped with the PX4Flow~\cite{honegger2013open} optical flow sensor, a single beam Lidar-lite v3 for altitude measurement, two $\unit[10]{W}$ LED light bars in both front arms, and four low-power LEDs looking down for providing additional illumination for the looking forward camera and the optical flow sensor respectively. Figure~\ref{fig:LTUplatfrom} presents the platform, highlighting it's dimensions and the sensor configuration, however the 2D lidar measurements are not used during the experimental trials.  

\begin{figure}[htbp!]
    \centering
\includegraphics[width=\linewidth]{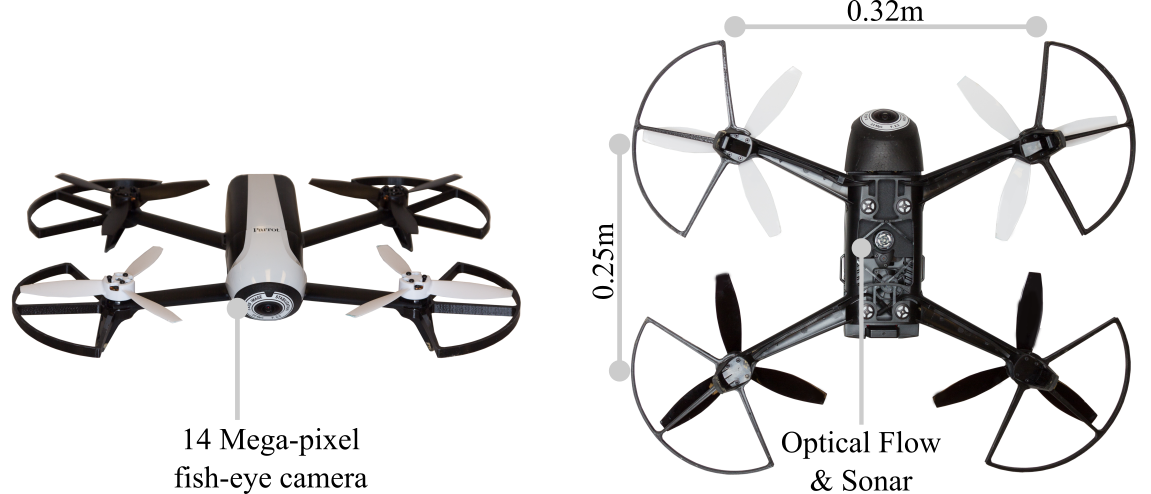}
    \caption{The commercial quad-copter Parrot Bebop 2.}
    \label{fig:bebop}
\end{figure}

The Parrot Bebop 2 weights $\unit[0.5]{kg}$, offering $\unit[25]{mins}$ of flight time and it is equipped with a forward looking camera,  a downward looking optical flow and sonar sensors as depicted in Figure~\ref{fig:bebop}. The platform provides a WiFi link and all the necessary computation was performed on the ground station computer with an Intel Core i7-6600U CPU, 2.6GHz and 8GB RAM. The Bebop-Autonomy~\footnote{\url{https://bebop-autonomy.readthedocs.io/}} is used for the estimation of the states and controlling the platform.
\subsection{Train and Experimental Evaluation of the CNN} 
The performance of the proposed \gls{cnn} method is evaluated under various conditions and cases, $\unit[790]{m}$ deep in an underground mine, without any natural illumination sources. The tunnel morphology resembled an \textit{S} shape environment with small inclination. The dimensions of the area were $6 (\text{width}) \times 4 (\text{height}) \times 300 (\text{length}) \unit{m^3}$.

\begin{figure}[htbp!]
  \centering
 \includegraphics[width=\linewidth]{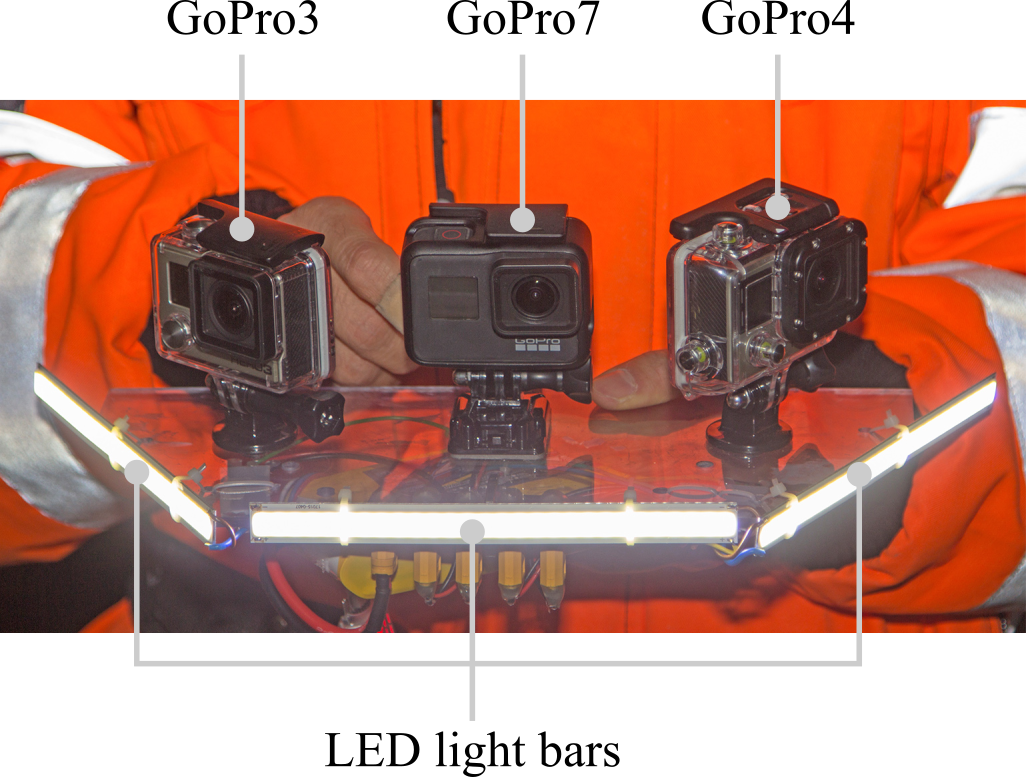}
      \caption{The triple camera setup for obtaining training data sets for \gls{cnn}.}
        \label{fig:triplecamerasetup}
\end{figure}

In order to train the \gls{cnn}, the setup, depicted in Figure~\ref{fig:triplecamerasetup}, was carried along a tunnel by a person, guaranteeing that the middle camera is looking towards the tunnel axis, and keeping the altitude fixed at $\unit[1]{m}$. To reduce dependency on a specific camera, three different cameras: 1) Gopro Hero 7, 2) Gopro Hero 4, and 3) GoPro Hero 3~\footnote{\url{https://gopro.com/}} are mounted with separated LED light bars pointing towards the field of view of each camera. The Gopro Hero 7, Gopro Hero 4, and GoPro Hero 3 record with $3840 \times 2160$ pixels at $\unit[60]{fps}$, $3840 \times 2160$ pixels at $\unit[30]{fps}$, and $1920 \times 1080$ pixels at $\unit[30]{fps}$ correspondingly. Moreover, the light bars are calibrated for providing equal illumination power and to consider uncertainties in the illumination level, the data sets are collected with different illumination levels. The collected videos are converted to images, while a down-sample to $5$~FPS is taking place to reduce the redundancy of the images, while all of them are labeled based on the direction of the camera. 

\begin{figure}[htbp!] \vspace{0.2cm}
    \centering
    \includegraphics[width=\linewidth]{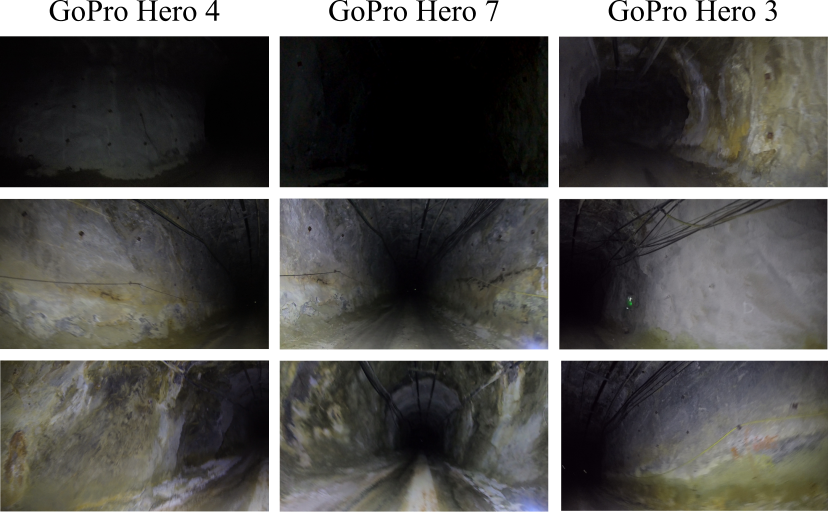}
    \caption{Examples of collected training data-sets from an underground mine. The left, center and right images are from cameras looking towards left, center and right correspondingly.}
    \label{fig:datasetscnn}
\end{figure}

Figure~\ref{fig:datasetscnn} shows the sample images from the underground mine environment, collected by the triple camera setup. Moreover the collected data-sets are from different tunnel areas than the tunnel where the \gls{mav} autonomous navigation was evaluated. In all cases the same trained network has been used. Each class of the \gls{cnn} has been trained with 1800 images corresponding to $\unit[80]{m}$ tunnel length.

\subsubsection{Effect of Sensor Variety}
In this case, the autonomous navigation of ROSflight based quad-copter and Parrot Bebop 2 with the \gls{cnn} module is evaluated. The ROSflight based quad-copter is equipped with a PlayStation Eye camera, which is operated at 20\,$\unit{fps}$ and with a resolution of $\unit[640 \times 480]{pixels}$. The Parrot Bebop 2 has 14 mega-pixels with fish-eye lens that provides $\unit[1920 \times 1080]{pixels}$ at $\unit[30]{fps}$. In both platforms, the desired altitude was $\unit[1]{m}$ with constant desired $v_{d,x}=\unit[0.1]{m/s}$ and $v_{d,y}=\unit[0.0]{m/s}$, while the LED light bars provide $\unit[460]{lux}$ illumination in $\unit[1]{m}$ distance. 

\begin{figure}[htbp!]
    \centering
    \includegraphics[width=\linewidth]{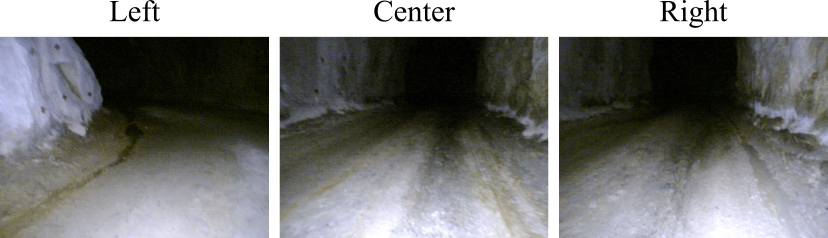}
    \caption{Image samples acquired from the PlayStation Eye camera. The class of images from the \gls{cnn} is depicted above the images.}
    \label{fig:cnnpixyimage} 
\end{figure}

The \gls{cnn} classifies the image with corresponding label and constant heading rate commands of $\unit[-0.2]{rad/s}$, $\unit[0.0]{rad/s}$, and $\unit[0.2]{rad/s}$ are generated for the \textit{left}, \textit{center} and \textit{right} labels respectively. In Figure~\ref{fig:cnnpixyimage} sample images from the PlayStation Eye camera are presented and the resulting class is depicted above each image.

\begin{figure}[!htb]
    \centering
\includegraphics[width=\linewidth]{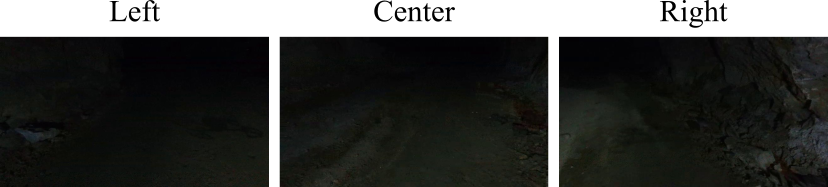}
    \caption{Examples of images from forward looking camera of the Parrot Bebop 2.}
    \label{fig:case1bebop}
\end{figure}

Furthermore, Figure~\ref{fig:case1bebop} depicts the sample images from Parrot Bebop 2 on-board camera, while the classes are presented again above each image. 

The trained network can provide accurate classification, while different types of visual sensors are used in the area that have not been included in the training data-sets.

\subsubsection{Effect of illumination}
\begin{figure}[!htb] \vspace{0.2cm}
    \centering
\includegraphics[width=\linewidth]{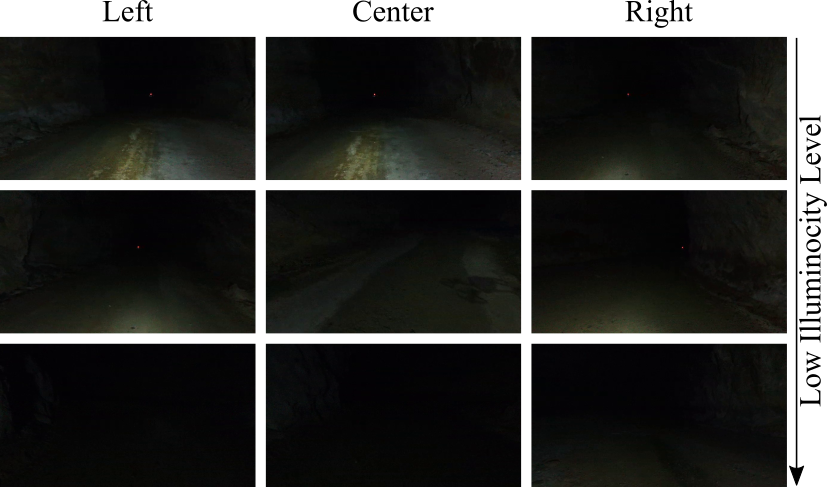}
    \caption{Image samples from the looking forward camera of the Parrot Bebop 2 with different illumination levels.}
    \label{fig:cnnpixyimage_bebop}
\end{figure}

In this case, the Parrot Bebop 2 performed autonomous navigation, while different level of illumination is provided and $z_d=\unit[1]{m}$, $v_{d,x}=\unit[0.1]{m/s}$, and $v_{d,y}=\unit[0.0]{m/s}$. In Figure~\ref{fig:cnnpixyimage_bebop} different images with different levels of illumination are shown. The \gls{cnn} provides correct heading rate commands, even though in some of the cases the images appear to be identical.

\subsubsection{Effect of Velocity}
The Parrot Bebop 2 is used for autonomous navigation in the mine with different velocities of $v_{d,x}=\{\unit[0.1, 0.5, 1]{m/s}\}$, constant altitude of $z_d=\unit[1]{m}$ and $v_{d,y}=\{\unit[0.0]{m/s}\}$. The increase in the velocity, reduces the quality of the images, however the \gls{cnn} provides correct heading rate commands for the \gls{mav} in all the evaluated velocities. 

\subsubsection{Evaluation in different mine Environment}
\begin{figure}[!htb]
    \centering
\includegraphics[width=\linewidth]{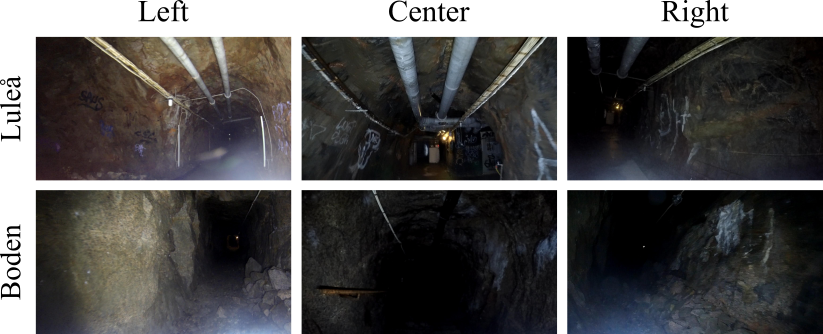}
\caption{Photos from an underground mining tunnel in Sweden where the experiments took place.}
    \label{fig:testdatasets}
\end{figure}

In order to test the general applicability of the trained \gls{cnn}, data-sets from~\cite{SinaRobio2018} are used to evaluate the performance of the method. The data-sets are labeled to three categories and the two underground tunnels are located in Lule\aa~~and Boden in Sweden with a different size dimension and structure when compared to the Boliden mine environment. Figure~\ref{fig:testdatasets} provides photos from two underground tunnels.

The trained \gls{cnn} provides a $78.4\%$ and $96.7\%$ accuracy score for Lule\aa~~and Boden underground tunnels respectively. The accuracy is calculated by the number of correct classified images over the total number of images. Moreover, Tables~\ref{table:confusionnmatrixboden} and~\ref{table:confusionnmatrixlulea} provide the confusion matrix for each underground tunnel.



\section{Conclusions} \label{Conclusions}
This article studies the problem of navigation in underground dark environments with resource constrained aerial platforms. The proposed method relies on velocities and heading rate commands and provides collision free navigation in unknown mine tunnels. The \gls{mav} is equipped with a looking forward camera and the \gls{cnn} method is used to classify the images from the camera to \textit{left}, \textit{center}, \textit{right}, while the heading rate command is generated based on the image labels. The framework has been evaluated in field trials performed in an underground mine in Sweden and has successfully managed to navigate in an \textit{S} shape environment autonomously and without collision to the wall surfaces. Moreover, the trained \gls{cnn} is evaluated with different visual sensors, velocities and illumination levels and in all these cases the \gls{cnn} provided correct heading rate commands. The trained \gls{cnn} was evaluated in two completely different underground tunnel data-sets and provided a high accuracy score in both cases. 
\begin{table}[htbp!]
\centering
\caption{The confusion matrix for the Lule\aa~~underground tunnel data-set.} 
\label{table:confusionnmatrixboden}
{\renewcommand{\arraystretch}{1.3}
\begin{tabular}{ccccc}
                              &                             & \multicolumn{3}{c}{Predicted Outcome}                                                 \\ \cline{3-5} 
\multirow{4}{*}{\rotatebox{90}{\parbox{1.5cm}{Actual \\ Class}}} & \multicolumn{1}{c|}{}       & \multicolumn{1}{c|}{left} & \multicolumn{1}{c|}{center} & \multicolumn{1}{c|}{right} \\ \cline{2-5} 
                              & \multicolumn{1}{|c|}{left}   & \multicolumn{1}{c|}{85.5\%}    & \multicolumn{1}{c|}{3.1\%}      & \multicolumn{1}{c|}{11.2\%}      \\ \cline{2-5} 
                              & \multicolumn{1}{|c|}{center} & \multicolumn{1}{c|}{19.2\%}    & \multicolumn{1}{c|}{66.8\%}      & \multicolumn{1}{c|}{13.9\%}      \\ \cline{2-5} 
                              & \multicolumn{1}{|c|}{right} & \multicolumn{1}{c|}{16.35\%}    & \multicolumn{1}{c|}{0.7\%}      & \multicolumn{1}{c|}{82.8\%}      \\ \cline{2-5} 
\end{tabular}
}
\end{table}

\begin{table}[htbp!]
\centering
\caption{The confusion matrix for the Boden underground tunnel data-set.}
\label{table:confusionnmatrixlulea}
{\renewcommand{\arraystretch}{1.3}
\begin{tabular}{ccccc}
                              &                             & \multicolumn{3}{c}{Predicted Outcome}                                                 \\ \cline{3-5} 
\multirow{4}{*}{\rotatebox{90}{\parbox{1.5cm}{Actual \\ Class}}} & \multicolumn{1}{c|}{}       & \multicolumn{1}{c|}{left} & \multicolumn{1}{c|}{center} & \multicolumn{1}{c|}{right} \\ \cline{2-5} 
                              & \multicolumn{1}{|c|}{left}   & \multicolumn{1}{c|}{97.6\%}    & \multicolumn{1}{c|}{2.4\%}      & \multicolumn{1}{c|}{0\%}      \\ \cline{2-5} 
                              & \multicolumn{1}{|c|}{center} & \multicolumn{1}{c|}{0.6\%}    & \multicolumn{1}{c|}{98.1\%}      & \multicolumn{1}{c|}{1.2\%}      \\ \cline{2-5} 
                              & \multicolumn{1}{|c|}{right} & \multicolumn{1}{c|}{2.4\%}    & \multicolumn{1}{c|}{2.4\%}      & \multicolumn{1}{c|}{95.0\%}      \\ \cline{2-5} 
\end{tabular}
}
\end{table}
\section{Acknowledgement} \label{Acknowledgement}
We would like to thank Boliden AB for providing the access to underground mine and for giving the opportunity to evaluate the proposed methods in real life challenging conditions and Wolfit AB for the valuable guidance and support during all our mine visits.
\bibliography{mybib}

\begin{thebibliography}{10}
\providecommand{\url}[1]{#1}
\csname url@samestyle\endcsname
\providecommand{\newblock}{\relax}
\providecommand{\bibinfo}[2]{#2}
\providecommand{\BIBentrySTDinterwordspacing}{\spaceskip=0pt\relax}
\providecommand{\BIBentryALTinterwordstretchfactor}{4}
\providecommand{\BIBentryALTinterwordspacing}{\spaceskip=\fontdimen2\font plus
\BIBentryALTinterwordstretchfactor\fontdimen3\font minus
  \fontdimen4\font\relax}
\providecommand{\BIBforeignlanguage}[2]{{%
\expandafter\ifx\csname l@#1\endcsname\relax
\typeout{** WARNING: IEEEtran.bst: No hyphenation pattern has been}%
\typeout{** loaded for the language `#1'. Using the pattern for}%
\typeout{** the default language instead.}%
\else
\language=\csname l@#1\endcsname
\fi
#2}}
\providecommand{\BIBdecl}{\relax}
\BIBdecl

\bibitem{kanellakis2018towards}
C.~Kanellakis, S.~S. Mansouri, G.~Georgoulas, and G.~Nikolakopoulos, ``{Towards
  Autonomous Surveying of Underground Mine Using MAVs},'' in
  \emph{International Conference on Robotics in Alpe-Adria Danube
  Region}.\hskip 1em plus 0.5em minus 0.4em\relax Springer, 2018, pp. 173--180.

\bibitem{rogers2017distributed}
J.~G. Rogers, R.~E. Sherrill, A.~Schang, S.~L. Meadows, E.~P. Cox, B.~Byrne,
  D.~G. Baran, J.~W. Curtis, and K.~M. Brink, ``Distributed subterranean
  exploration and mapping with teams of uavs,'' in \emph{Ground/Air Multisensor
  Interoperability, Integration, and Networking for Persistent ISR VIII}, vol.
  10190.\hskip 1em plus 0.5em minus 0.4em\relax International Society for
  Optics and Photonics, 2017, p. 1019017.

\bibitem{mansouri2018cooperative}
S.~S. Mansouri, C.~Kanellakis, E.~Fresk, D.~Kominiak, and G.~Nikolakopoulos,
  ``Cooperative coverage path planning for visual inspection,'' \emph{Control
  Engineering Practice}, vol.~74, pp. 118--131, 2018.

\bibitem{mansouri20182d}
S.~S. Mansouri, C.~Kanellakis, G.~Georgoulas, D.~Kominiak, T.~Gustafsson, and
  G.~Nikolakopoulos, ``2d visual area coverage and path planning coupled with
  camera footprints,'' \emph{Control Engineering Practice}, vol.~75, pp. 1--16,
  2018.

\bibitem{matthaei2013swarm}
J.~Matthaei, T.~Kr{\"u}ger, S.~Nowak, and U.~Bestmann, ``Swarm exploration of
  unknown areas on mars using slam,'' in \emph{International Micro Air Vehicle
  Conference and Flight Competition (IMAV)}, 2013.

\bibitem{ozaslan2017autonomous}
T.~{\"O}zaslan, G.~Loianno, J.~Keller, C.~J. Taylor, V.~Kumar, J.~M.
  Wozencraft, and T.~Hood, ``Autonomous navigation and mapping for inspection
  of penstocks and tunnels with {MAVs},'' \emph{IEEE Robotics and Automation
  Letters}, vol.~2, no.~3, pp. 1740--1747, 2017.

\bibitem{krizhevsky2012imagenet}
A.~Krizhevsky, I.~Sutskever, and G.~E. Hinton, ``Imagenet classification with
  deep convolutional neural networks,'' in \emph{Advances in neural information
  processing systems}, 2012, pp. 1097--1105.

\bibitem{schmid2014autonomous}
K.~Schmid, P.~Lutz, T.~Tomi{\'c}, E.~Mair, and H.~Hirschm{\"u}ller,
  ``Autonomous vision-based micro air vehicle for indoor and outdoor
  navigation,'' \emph{Journal of Field Robotics}, vol.~31, no.~4, pp. 537--570,
  2014.

\bibitem{gohl2014towards}
P.~Gohl, M.~Burri, S.~Omari, J.~Rehder, J.~Nikolic, M.~Achtelik, and
  R.~Siegwart, ``Towards autonomous mine inspection,'' in \emph{Applied
  Robotics for the Power Industry (CARPI), 2014 3rd International Conference
  on}.\hskip 1em plus 0.5em minus 0.4em\relax IEEE, 2014, pp. 1--6.

\bibitem{mansouri2019autonomouscontour}
S.~S. Mansouri, M.~Casta{\~n}o, C.~Kanellakis, and G.~Nikolakopoulos,
  ``Autonomous mav navigation in underground mines using darkness contours
  detection,'' in \emph{International Conference on Computer Vision
  Systems}.\hskip 1em plus 0.5em minus 0.4em\relax Springer, 2019, pp.
  164--174.

\bibitem{adhikari2018accurate}
S.~P. Adhikari, C.~Yang, K.~Slot, and H.~Kim, ``Accurate natural trail
  detection using a combination of a deep neural network and dynamic
  programming,'' \emph{Sensors}, vol.~18, no.~1, p. 178, 2018.

\bibitem{giusti2016machine}
A.~Giusti, J.~Guzzi, D.~C. Ciresan, F.-L. He, J.~P. Rodr{\'\i}guez, F.~Fontana,
  M.~Faessler, C.~Forster, J.~Schmidhuber, G.~Di~Caro \emph{et~al.}, ``A
  machine learning approach to visual perception of forest trails for mobile
  robots.'' \emph{IEEE Robotics and Automation Letters}, vol.~1, no.~2, pp.
  661--667, 2016.

\bibitem{ran2017convolutional}
L.~Ran, Y.~Zhang, Q.~Zhang, and T.~Yang, ``Convolutional neural network-based
  robot navigation using uncalibrated spherical images,'' \emph{Sensors},
  vol.~17, no.~6, p. 1341, 2017.

\bibitem{smolyanskiy2017toward}
N.~Smolyanskiy, A.~Kamenev, J.~Smith, and S.~Birchfield, ``Toward low-flying
  autonomous {MAV} trail navigation using deep neural networks for
  environmental awareness,'' \emph{arXiv preprint arXiv:1705.02550}, 2017.

\bibitem{kourislearning}
A.~Kouris and C.-S. Bouganis, ``Learning to fly by myself: A self-supervised
  cnn-based approach for autonomous navigation,'' in \emph{Intelligent Robots
  and Systems (IROS), 2018 IEEE/RSJ International Conference on}, 2017.

\bibitem{SinaRobio2018}
S.~S. Mansouri, C.~Kanellakis, G.~Georgoulas, and G.~Nikolakopoulos, ``Towards
  {MAV} navigation in underground mine using deep learning,'' in \emph{IEEE
  International Conference on Robotics and Biomimetics (ROBIO)}, 2018.

\bibitem{NIKOLAKOPOULOS201566}
G.~Nikolakopoulos, T.~Gustafsson, P.~Martinsson, and U.~Andersson, ``A vision
  of zero entry production areas in mines∗∗this work has been partially
  funded by the sustainable mining and innovation for the future research
  program.'' \emph{IFAC-PapersOnLine}, vol.~48, no.~17, pp. 66 -- 68, 2015, 4th
  IFAC Workshop on Mining, Mineral and Metal Processing MMM 2015.

\bibitem{gade2016seven}
K.~Gade, ``The seven ways to find heading,'' \emph{The Journal of Navigation},
  vol.~69, no.~5, pp. 955--970, 2016.

\bibitem{cirecsan2012multi}
D.~Cire{\c{s}}an, U.~Meier, and J.~Schmidhuber, ``Multi-column deep neural
  networks for image classification,'' \emph{arXiv preprint arXiv:1202.2745},
  2012.

\bibitem{bui2016using}
H.~M. Bui, M.~Lech, E.~Cheng, K.~Neville, and I.~S. Burnett, ``Using grayscale
  images for object recognition with convolutional-recursive neural network,''
  in \emph{IEEE Sixth International Conference on Communications and
  Electronics (ICCE)}, 2016, pp. 321--325.

\bibitem{chollet2015keras}
F.~Chollet \emph{et~al.}, ``Keras,'' \url{https://github.com/fchollet/keras},
  2015.

\bibitem{kingma2014adam}
D.~P. Kingma and J.~Ba, ``Adam: A method for stochastic optimization,''
  \emph{arXiv preprint arXiv:1412.6980}, 2014.

\bibitem{jackson2016rosflight}
J.~{Jackson}, G.~{Ellingson}, and T.~{McLain}, ``{ROSflight: A lightweight,
  inexpensive MAV research and development tool},'' in \emph{2016 International
  Conference on Unmanned Aircraft Systems (ICUAS)}, June 2016, pp. 758--762.

\bibitem{parrot2016parrot}
S.~Parrot, ``Parrot bebop 2,'' \emph{Retrieved from Parrot. com: http://www.
  parrot. com/products/bebop2}, 2016.

\bibitem{quigley2009ros}
M.~Quigley, K.~Conley, B.~Gerkey, J.~Faust, T.~Foote, J.~Leibs, R.~Wheeler, and
  A.~Y. Ng, ``{ROS}: an open-source robot operating system,'' in \emph{ICRA
  workshop on open source software}, vol.~3, no. 3.2.\hskip 1em plus 0.5em
  minus 0.4em\relax Kobe, Japan, 2009, p.~5.

\bibitem{honegger2013open}
D.~Honegger, L.~Meier, P.~Tanskanen, and M.~Pollefeys, ``An open source and
  open hardware embedded metric optical flow cmos camera for indoor and outdoor
  applications,'' in \emph{Robotics and Automation (ICRA), 2013 IEEE
  International Conference on}.\hskip 1em plus 0.5em minus 0.4em\relax IEEE,
  2013, pp. 1736--1741.

\end{thebibliography}
\end{document}